\documentclass{article}

\usepackage{microtype}
\usepackage{graphicx}
\usepackage{multirow}
\usepackage{booktabs} 
\usepackage{tikz}

\usepackage{hyperref}
\usepackage{amsmath} 
\usepackage{amssymb}  
\usepackage{subcaption}
\newcommand{\KL}{D_{\mathrm{KL}}}

\DeclareGraphicsExtensions{.png,.pdf}

\usepackage[accepted]{icml2019}

\icmltitlerunning{Diffusion Variational Autoencoders}

\newcommand{\DeltaVAE}{$\Delta\!$VAE}

\begin{document}

\twocolumn[
\icmltitle{Diffusion Variational Autoencoders}



\icmlsetsymbol{equal}{*}

\begin{icmlauthorlist}
\icmlauthor{Luis A. P\'{e}rez Rey}{equal,to}
\icmlauthor{Vlado Menkovski}{equal,to}
\icmlauthor{Jacobus W. Portegies}{equal,to}
\end{icmlauthorlist}

\icmlaffiliation{to}{Eindhoven University of Technology, Eindhoven, The Netherlands}

\icmlcorrespondingauthor{Luis A. P\'{e}rez Rey}{l.a.perez@tue.nl}
\icmlcorrespondingauthor{Jacobus W. Portegies}{j.w.portegies@tue.nl}

\icmlkeywords{Machine Learning, ICML}

\vskip 0.3in
]



\printAffiliationsAndNotice{\icmlEqualContribution} 

\begin{abstract}

A standard Variational Autoencoder, with a Euclidean latent space, is structurally incapable of capturing topological properties of certain datasets. To remove topological obstructions, we introduce Diffusion Variational Autoencoders with \emph{arbitrary} manifolds as a latent space. A Diffusion Variational Autoencoder uses transition kernels of Brownian motion on the manifold. In particular, it uses properties of the Brownian motion to implement the reparametrization trick and fast approximations to the KL divergence. 

We show that the Diffusion Variational Autoencoder is capable of capturing topological properties of synthetic datasets. Additionally, we train MNIST on spheres, tori, projective spaces, $SO(3)$, and a torus embedded in $\mathbb{R}^3$. Although a natural dataset like MNIST does not have latent variables with a clear-cut topological structure, training it on a manifold can still highlight topological and geometrical properties.
\end{abstract}

\section{Introduction}
\label{introduction}

A large part of unsupervised learning is devoted to the extraction of meaningful latent factors that explain a certain data set. The terminology around Variational Autoencoders suggests that they are a good tool for this task. 

A Variational Autoencoder \cite{Kingma2014, Rezende2014StochasticModels} consists of a ``latent space" $Z$ (often just Euclidean space), a probability measure $\mathbb{P}_Z$ on $Z$, an encoder from the data space $X$ to $Z$ and a decoder from $Z$ to $X$. The term ``latent space" suggests that an element in $z$ captures semantic information about its decoded data point $x = \mathsf{Dec}(z)$. But is this interpretation actually warranted? 
Especially considering that in practice, a Variational Autoencoder is often trained without any knowledge about an underlying generative process. The only available information is the dataset itself. 
Nonetheless, one may train a Variational Autoencoder with any ``latent space" $Z$ and purely based on this terminology, one could be tempted to think of elements in $Z$ as latent variables and factors explaining the data. 

\begin{figure}[t]
\centering
\begin{minipage}{0.49\columnwidth}
  \centering
\includegraphics[width=\textwidth]{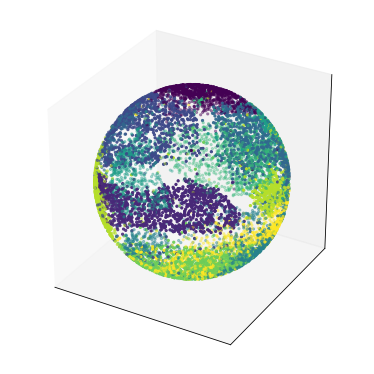}
\label{fig:1a}
\end{minipage}
\begin{minipage}{0.49\columnwidth}
  \centering
\includegraphics[width=\textwidth]{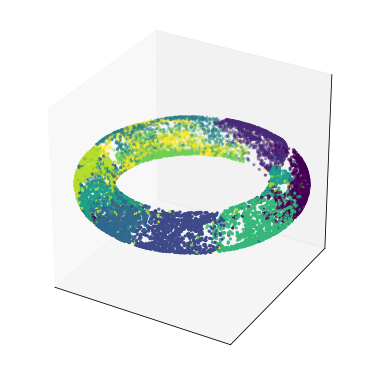}
\label{fig:1a}
\end{minipage}
\begin{minipage}{0.5\textwidth}
  \centering
\includegraphics[width=0.45\textwidth]{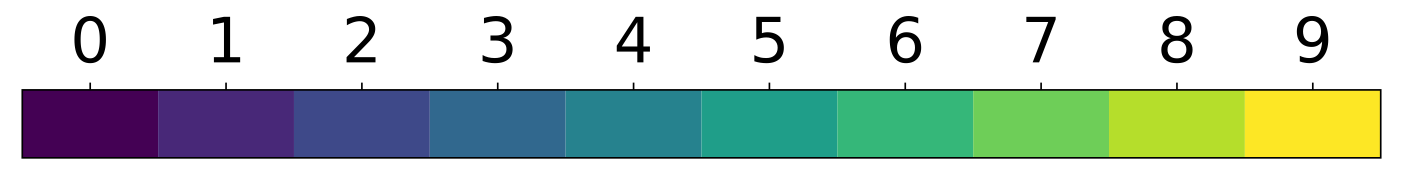}
\end{minipage}
\caption{MNIST trained on a sphere and on a torus}
\label{fig:sphere-torus-latent}
\end{figure}

Certainly, there is a heuristic argument that gives a partial justification of this interpretation. The starting point and guiding principle is that of Occam's razor, that such latent factors might arise from the construction of simple, low-complexity models that explain the data \cite{portegies2018ergo}. Strict formalizations of Occam's razor exist, through Kolmogorov complexity and inductive inference \cite{Solomonoff1964AI,Schmidhuber1997DiscoveringCapability}, but are often computationally intractable. A more practical approach leads to the principle of minimum description length \cite{Rissanen1978ModelingDescription, hinton1994autoencoders} and variational inference \cite{Honkela2004VariationalLearning}.

From a different angle, part of the interpretation as latent space is warranted by the loss function of the Variational Autoencoder, which stimulates a continuous dependence between the latent variables and the corresponding data points. Close-by points in data space should also be close-by in latent space. This suggests that a Variational Autoencoder could capture topological and geometrical properties of a data set. 

However, a standard Variational Autoencoder (with a Euclidean latent space) is at times structurally incapable of accurately capturing topological properties of a data set. Take for example the case of a spinning object placed on a turntable and being recorded by a camera from a fixed position. The data set for this example is the collection of all frames. 
The true latent factor is the angle of the turntable. However, the space of angles is topologically and geometrically different from Euclidean space. In an extreme example, if we train a Variational Autoencoder with a one-dimensional latent space on the pictures from the object on the turntable, there will be pictures taken from almost the same angle ending up at completely different parts of the latent space.

This phenomenon has been called manifold mismatch \cite{Davidson2018, Falorsi2018ExplorationsAuto-Encoding}. 
To match the latent space with the data structure, Davidson et~al. implemented spheres as latent spaces, whereas Falorsi  et~al. implemented the special orthogonal group $SO(3)$. 

As further examples of datasets with topologically nontrivial latent factors, we can think of many translations of the same periodic picture, where the translation is the latent variable, or many pictures of the same object which has been rotated arbitrarily. In these cases, there are still clear latent variables, but their topological and geometrical structure is neither that of Euclidean space nor that of a sphere, but rather that of a torus and that of the $SO(3)$ respectively.

To address the problem of manifold mismatch, we developed the Diffusion Variational Autoencoder (\DeltaVAE) which allows for an \emph{arbitrary} manifold as a latent space. Our implementation includes a version of the reparametrization trick, and a fast approximation of the evaluation of the KL divergence in the loss.

We implemented {\DeltaVAE}s with latent spaces of $d$-dimensional spheres, a two-dimensional flat torus, an embedded torus in $\mathbb{R}^3$, the special orthogonal group $SO(3)$ and the real projective spaces $\mathbb{RP}^d$. 

We trained a synthetic data set of translated images on a flat torus. Our results show that the {\DeltaVAE} can capture the topological properties of the data. We further observed that the success rate of the {\DeltaVAE} in capturing global topological properties depends on the weight of higher Fourier components in the image. In data sets with more pronounced lower Fourier components the {\DeltaVAE} is more successful in capturing the topology. 

Mainly as a proof of concept, we trained on MNIST using the manifolds mentioned above. 

\section{Related work}

Our work originated out of the search for algorithms that find semantically meaningful latent factors of data. The use of VAEs and their extensions to this end has mostly taken place in the context of \emph{disentanglement of latent factors} \cite{Higgins2016,Higgins2018TowardsRepresentations,Burgess2018Understandingbeta-VAE}. Examples of extensions that aim at disentangling latent factors are the $\beta$-VAE \cite{Higgins2016}, the factor-VAE \cite{Kim2018}, the $\beta$-TCVAE \cite{Chen2018IsolatingAutoencoders} and the DIP-VAE \cite{kumar2018variational}. 

However, the examples in the introduction already show that in some situations, the topological structure of the latent space makes it practically impossible to disentangle latent factors. The latent factors are inherently, topologically entangled: in the case of a 3d rotation of an object, one cannot assign globally linearly independent angles of rotation. 

Still, it is exactly global topological properties that we feel a VAE has a chance of capturing. What do we mean by this? One instance of `capturing' topological structure is when the encoder and decoder of the VAE provide bijective, continuous maps between data and latent space, also called homeomorphic auto-encoding \cite{Falorsi2018ExplorationsAuto-Encoding, deHaan2018TopologicalAuto-Encoding}. This can only be done when the latent space has a particular topological structure, for instance that of a particular manifold.

One of the main challenges when implementing a manifold as a latent space is the design of the reparametrization trick. In \cite{Davidson2018}, a VAE was implemented with a hyperspherical latent space. To our understanding, they implemented a reparametrization function which was discontinuous; see also Section \ref{se:reparametrization-trick} below.

If a manifold has the additional structure of a Lie group, this structure allows for a more straightforward implementation of the reparametrization trick  \cite{Falorsi2018ExplorationsAuto-Encoding}. 
In our work, we do not assume the additional structure of a Lie group, but develop a reparametrization trick that works for general submanifolds of Euclidean space, and therefore by the Whitney (respectively Nash) embedding theorem, for general closed (Riemannian) manifolds. 

The  method that we use has similarities with the approach of Hamiltonian Variational Inference \cite{Salimans2014MarkovGap}. Moreover, the implementation of a manifold as a latent space can be seen as enabling a particular, informative, prior distribution. In that sense, our work relates to \cite{Dilokthanakul2016DeepAutoencoders, Tomczak2017VAEVampPrior}. The prior distribution we implement is very degenerate, in that it is does not assign weight to points outside of the manifold.

There are also other ways to implement approximate Bayesian inference on Riemannian manifolds. For instance, Liu and Zhu adapted the Stein variational gradient method to enable training on a Riemannian manifold \cite{Liu2017RiemannianInference}. However, their proposed method is rather expensive computationally.

The family of approximate posteriors that we implement is a direct generalization of the standard choice for a Euclidean VAE. Indeed, the Gaussian distributions are solutions to the heat equations, i.e. they are transition kernels of Brownian motion. One may want to increase the flexibility of the family of approximate posterior distributions, for instance by applying normalizing flows \cite{rezende2015variational,Kingma2016ImprovingFlow,Gemici2016NormalizingManifolds}.

\section{Methods}
\subsection{Variational autoencoders}

A VAE has generally the following ingredients:
{\begin{itemize}
\setlength{\itemsep}{0em}
    \item a latent space $Z$,
    \item a prior probability distribution $\mathbb{P}_Z$ on $Z$,
    \item a family of encoder distributions $\mathbb{Q}_Z^\alpha$ on $Z$, parametrized by $\alpha$ in a parameter space $\mathcal{A}$, 
    \item a family of decoder distributions $\mathbb{P}_X^\beta$ on data space $X$, parametrized by $\beta$ in a parameter space $\mathcal{B}$; in the usual setup, and in our paper, in fact $\mathcal{B}$ corresponds to the data space, and refers to the mean of a Gaussian distribution with identity covariance,
    \item an encoder neural network $\boldsymbol{\alpha}$ which maps from data space $X$ to the parameter space $\mathcal{A}$,
    \item a decoder neural network $\boldsymbol{\beta}$ which maps from latent space $Z$ to parameter space $\mathcal{B}$. 
\end{itemize}
}
The weights of these neural networks are optimized as to minimize the negated evidence lower bound (ELBO) 
\[
    \mathcal{L}(x) = -\mathbb{E}_{ z\sim\mathbb{Q}_{Z}^{\boldsymbol{\alpha}(x)}}\left[\log{p_X^{\boldsymbol{\beta}(z)}(x)}\right]\\
    + \KL\left(\mathbb{Q}_{Z}^{\boldsymbol{\alpha}(x)}||\mathbb{P}_Z\right).
\]
The first term is called \emph{reconstruction error} (RE); up to additive and multiplicative constants it equals the mean squared error (MSE). The second term is called the KL-loss.

In a very common implementation, both latent space $Z$ and data space $X$ are Euclidean, and the families of decoder and encoder distributions are multivariate Gaussian. The encoder and decoder networks then assign to a datapoint or a latent variable a mean and a variance respectively.

When we implement $Z$ as a Riemannian manifold, we need to find an appropriate prior distribution, for which we will choose the normalized Riemannian volume measure, a family of encoder distributions $\mathbb{Q}^\alpha_Z$, for which we will take transition kernels of Brownian motion, and an encoder network mapping to the correct parameters.

\subsection{Brownian motion on a Riemannian manifold}

We will briefly discuss Brownian motion on a Riemannian manifold, recommending lecture notes by Hsu \cite{Hsu2008} as a more extensive introduction.

In the paper, we always assume that $Z$ is a smooth Riemannian submanifold of Euclidean space, which is closed, i.e. it is compact and has no boundary. 

There are many different, equivalent definitions of Brownian motion. We present here the definition that is closest to our eventual approximation and implementation.

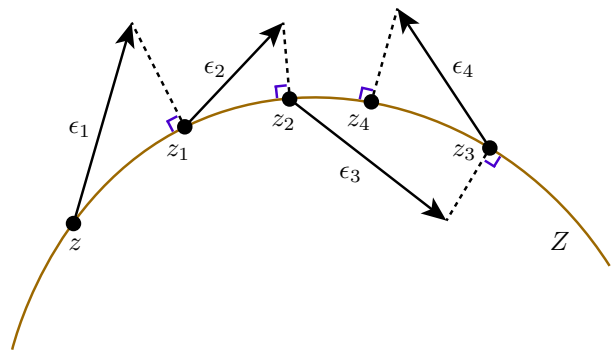
\begin{figure}[h]
\begin{tikzpicture}[      
        every node/.style={anchor=south west,inner sep=0pt},
        x=1mm, y=1mm,
      ]   
     \node (fig1) at (0,0)
       {\includegraphics[width=\columnwidth]{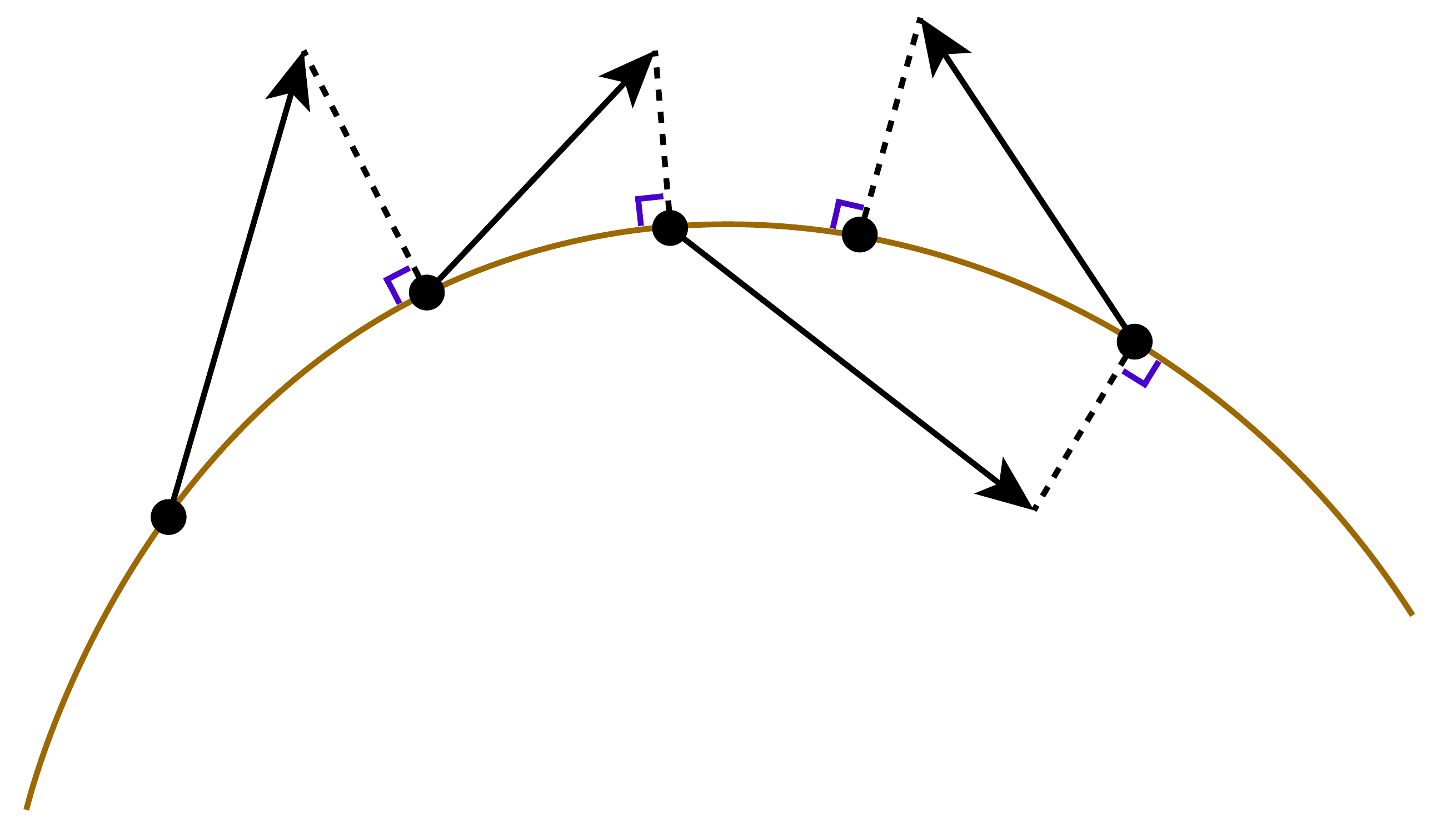}};
     \node (epsilon1) at (9,30) {$\epsilon_1$};
     \node (epsilon2) at (27,38) {$\epsilon_2$};
     \node (epsilon3) at (45,24) {$\epsilon_3$};
     \node (epsilon4) at (60,38) {$\epsilon_4$};
     \node (z) at (9,15) {$z$};
     \node (z1) at (22,27) {$z_1$};
     \node (z2) at (36,31) {$z_2$};
     \node (z3) at (60,27) {$z_3$};
     \node (z4) at (46,31) {$z_4$};
     \node (Z) at (73,15) {$Z$};
\end{tikzpicture}
    \caption{Random walk on a (one-dimensional) submanifold $Z$ of $\mathbb{R}^2$, with time step $\tau =1$.}
    \label{fig:random_walk}
\end{figure}

We will construct Brownian motion out of random walks on a manifold. We first fix a small time step $\tau > 0$. We will imagine a particle, jumping from point to point on the manifold after each time step, see also Fig.~\ref{fig:random_walk}. It will start off at a point $z \in Z$. We describe the first jump, after which the process just repeats. After time $\tau$, the particle makes a random jump $\sqrt{\tau} \epsilon_1$ from its current position, into the surrounding space, where $\epsilon_1$ is distributed according to a radially symmetric distribution in $\mathbb{R}^n$ with identity covariance matrix. The position of the particle after the jump, $z + \sqrt{\tau}\epsilon_1$, will therefore in general not be on the manifold, so we project the particle back: The particle's new position will be 
\[
z_1 = P( z + \sqrt{\tau} \epsilon_1 )
\]
where the closest-point-projection $P: \mathbb{R}^n \to Z$ assigns to every point $x \in \mathbb{R}^n$ the point in $Z$ that is closest to $x$. After another time $\tau>0$ the particle makes a new, independent, jump $\epsilon_2$ according to the same radially symmetric distribution, and its new position will be $z_2 = P(P(z + \sqrt{\tau} \epsilon_1) + \sqrt{\tau} \epsilon_2)$. This process just repeats.

Key to this construction, and also to our implementation, is the projection map $P$. It has nice properties, that follow from general theory of smooth manifolds. In particular, $P(x)$ smoothly depends on $x$, as long as $x$ is not too far away from $Z$.

This way, for $\tau>0$ fixed, we have constructed a random walk, a random path on the manifold. 
We can think of this path as a discretized version of Brownian motion. Let now $\tau_N$ be a sequence converging to $0$ as $N \to \infty$. For fixed $N \in \mathbb{N}$, we can construct a random walk with time step $\tau_N$, and get a random path $W^N: [0,\infty) \to Z$. 

The random paths $W^N$ converge as $N \to \infty$ to a random path $W$ (in distribution). This random path $W$ is called Brownian motion. The convergence statement can be made precise by for instance combining powerful, general results by \cite{Jrgensen1975} with standard facts from Riemannian geometry. But, because Riemannian manifolds are locally, i.e.~when you zoom in far enough, very similar to Euclidean space, the convergence result essentially comes down to the central limit theorem and its upgraded version, Donsker's invariance theorem.

In fact, $W$ can be interpreted as a Markov process, and even as a diffusion process. If $A$ is a subset of $Z$, the probability that the Brownian motion $W(t)$ started at $z$ is in the set $A$ at time $t$ is measured by a probability measure $\mathbb{Q}_Z^{t,z}$ applied to the set $A$. We denote the density of this measure with respect to the standard Riemannian volume measure by $q_Z(t;z,\cdot)$. The function $q_Z$ is sometimes referred to as the heat kernel. 

Let us close this subsection with an alternative description of the function $q_Z$. It is also characterized by the fact that for every function $u_0: Z \to \mathbb{R}$, the solution to the partial differential equation
\[
\begin{cases}
    \partial_t u = \tfrac{1}{2} \Delta u & \text{on } (0,\infty) \times Z\\
    u(t=0,\cdot)   = u_0 & \text{on } Z
\end{cases}
\]
is given by
\[
u(t,z) = \int_Z u_0(y) q_Z(t; z, y) dy.
\]

\subsection{Riemannian manifold as latent space}

A {\DeltaVAE} is a VAE with a Riemannian submanifold of Euclidean space as a latent space, and the transition probability measures of Brownian motion $\mathbb{Q}_Z^{t,z}$ as a parametric family of encoder distributions.
We propose the uniform distribution for $\mathbb{P}_Z$, which is the normalized standard measure on a Riemannian manifold (although the choice of prior distribution could easily be generalized). 

As in the standard VAE, we then implement functions $\mathbf{z}: X \to Z$ and $\mathbf{t} : X \to (0,\infty)$ as neural networks.

\subsection{ELBO}
We optimize the weights in the network, aiming to minimize the average loss for the loss function
\[
- \mathbb{E}_{z \sim \mathbb{Q}_Z^{\mathbf{t}(x), \mathbf{z}(x)}}\left[ \log p_X^{\boldsymbol{\beta}(z)}(x) \right]
+ \KL\left( \mathbb{Q}_{Z}^{\mathbf{t}(x), \mathbf{z}(x)} \| \mathbb{P}_Z \right).
\]
The first integral can often only be approached by sampling, and in that case it is often advantageous to perform a change of variables, commonly known as the \emph{reparametrization trick} \cite{Kingma2014}.

\subsection{Reparametrization trick}
\label{se:reparametrization-trick}

The reparametrization trick requires a space $\Gamma$ of noise variables, distributed according to $\mathbb{P}_\Gamma$, and a function $g:\Gamma \times (0,\infty) \times Z$ such that $g(\gamma, t, z)$ is approximated according to $\mathbb{Q}_Z^{t,z}$. 

The following example illustrates that finding such a function $g$ can be a nontrivial task, even for a relatively simple case where $Z= S^2$, the two-dimensional unit sphere in $\mathbb{R}^3$. In an attempt to construct a random variable distributed according to $\mathbb{Q}_{S^2}^{t,z}$, we may first sample a random tangent vector $\gamma$ at the south pole according to an appropriate distribution, and then walk along the great circle in that direction for a distance according to the length of the vector. In other words, the new point is obtained from applying the exponential map at the south pole to the random sample. 
Next, we compose with a rotation of the sphere that brings the south pole to the point $z$ on the sphere. More precisely, we select a map of rotations $T: S^2 \to SO(3)$ such that every $y \in S^2$ is the image of the south pole under the rotation $T(y)$. Then, a proposed reparametrization $g$ is given by
\[
g(\gamma, z) = T ( z ) \big[\exp_{\mathsf{SP}}( \gamma )\big].
\]

In some sense, this reparametrization works: One could find a distribution $\mathbb{P}_\Gamma$ on the tangent space at the south pole such that the distribution of $g(\gamma,z)$ is $\mathbb{Q}_{S^2}^{t, z}$. However, for topological reasons, in particular by the hairy ball theorem, it is impossible to find a \emph{continuous} map $T$ with this property. 

\subsection{Approximate reparametrization by random walk}
\label{se:reparametrization-rw}

Instead, we construct an \emph{approximate} reparametrization map by approximating Brownian motion by a random walk, similar to how we defined it in this paper. Starting from a point $z$ on the manifold, we set a random step in \emph{ambient space} $\mathbb{R}^n$. We then project back to the manifold and repeat: we take a new step and project back to the manifold. In total, we take $N$ steps, see Fig.~\ref{fig:random_walk}.

We define the function $g: \mathcal{E}^N \times (0,\infty) \times Z \to Z$ by
\[
\begin{split}
&g( \epsilon_1, \dots, \epsilon_N , t , z ) =\\
&P\left( \cdots P \left(P\left(z + \sqrt{\tfrac{t}{N}}\epsilon_1\right) + \sqrt{\tfrac{t}{N}}\epsilon_2 \right) \cdots + \sqrt{\tfrac{t}{N}} \epsilon_N \right).
\end{split}
\]
If we take $\epsilon_1, \dots, \epsilon_N$ as i.i.d.~random variables, distributed according to a radially symmetric distribution, then $y = g(\epsilon_1, \dots, \epsilon_N, z)$ is approximately distributed as a random variable with density $q_Z(t; z, \cdot )$. This approximation is very accurate for small times $t$, even for small values of $N$, if we take $\epsilon_1, \dots, \epsilon_N$ approximately Gaussian. The observation that for small times, the diffusion kernel $q_Z$ is approximately Gaussian, is also very helpful in approximating the KL term in the loss.

\subsection{Approximation of the KL-divergence}

Unlike the standard VAE, or the hyperspherical VAE with the Von-Mises distribution, the KL-term cannot be computed exactly for the \DeltaVAE. There are several techniques one could use to get, nonetheless, a good approximation of the KL divergence. We have implemented an asymptotic approximation, which we will describe first.

\subsubsection*{Asymptotic approximation}

We can use short-term asymptotics, i.e. a parametrix expansion, of the heat kernel on Riemannian manifolds to obtain asymptotic expansions of the entropy.

For a $d$-dimensional sphere of radius $R$, the scalar curvature $\mathsf{Sc}$ equals $d(d-1) /R^2$. By using a parametrix expansion of the heat kernel, see for instance \cite{Zhao2017ExactSVM}, we find that
\[
\begin{split}
&q_{S^d}(t;z,y) = \frac{1}{(2 \pi t)^{d/2}}\exp\left(-\frac{r^2}{2t}\right) \times \left(\frac{r}{\sin{r}}\right)^{\frac{d-1}{2}} \\
&\, \left(1 + \frac{\mathsf{Sc}\, t}{8 d r^2} [3 - d + (d-1)r^2 + (d-3) r \cot r] + O(t^2)\right)
\end{split}
\]
where $r$ is the geodesic distance between $y$ and $z$. 

In Appendix \ref{app:asymptotic_expansion}, we derive the following asymptotic behavior for the KL divergence for arbitrary Riemannian manifolds
\[
\begin{split}
\KL(\mathbb{Q}^{t,z} \| \mathbb{P}_Z )
&= \int_{Z} q_Z(t; z, y) \log q_Z(t;z,y) dy  \\&\qquad + \log \mathrm{Vol}(Z) \\
&= - \frac{d}{2} \log(2 \pi t) - \frac{d}{2} + \log \mathrm{Vol}(Z) \\
&\qquad  + \frac{1}{4} \mathsf{Sc}\, t + o(t).
\end{split}
\]
In our implementation, we restrict $t$ so that it cannot become too large, thus ensuring a certain accuracy of the asymptotic expansion.

\begin{figure}[!b]
\centering
\begin{minipage}{0.49\columnwidth}
  \centering
\includegraphics[width=\textwidth]{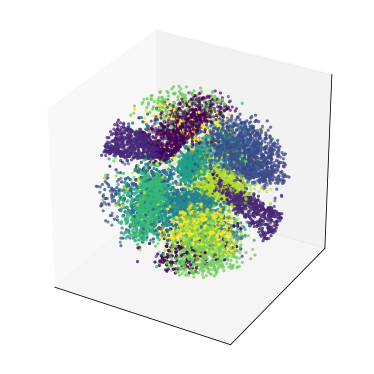}
\end{minipage}
\begin{minipage}{0.49\columnwidth}
  \centering
\includegraphics[width=\textwidth]{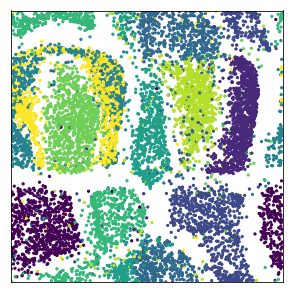}
\end{minipage}
\\
\begin{minipage}{0.49\columnwidth}
  \centering
\includegraphics[width=\textwidth]{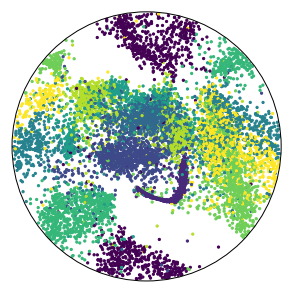}
\end{minipage}
\begin{minipage}{0.49\columnwidth}
  \centering
\includegraphics[width=\textwidth]{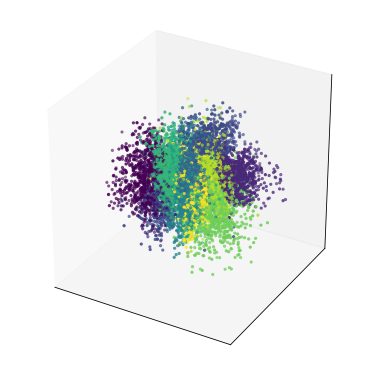}
\end{minipage}

\begin{minipage}{\columnwidth}
  \centering
\includegraphics[width=0.45\textwidth]{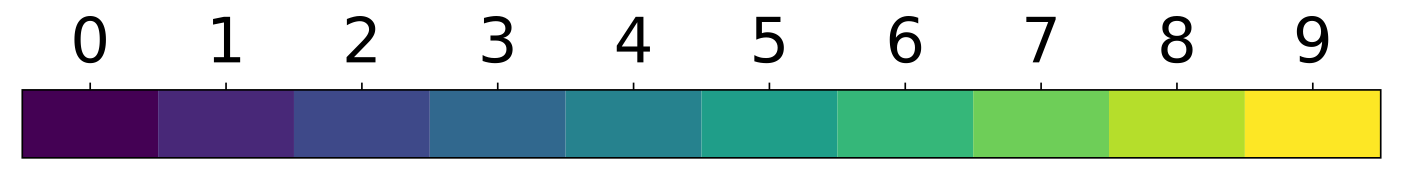}
\end{minipage}

\caption{Latent space representation of MNIST for $SO(3) \cong \mathbb{RP}^3$ (top-left), $\mathbb{RP}^2$ (bottom-left), flat torus (top-right), and $\mathbb{R}^3$ (bottom-right). The flat torus is represented as a square with periodic boundary conditions. The projective spaces are represented by a $3$- and $2$-dimensional ball respectively, for which every point on the boundary is identified with its reflection through the center. The effect of this identification can be seen, since the same digits that map close to a point on the boundary also map close to the reflected point.}
\label{fig:mnist-latent-rest}
\end{figure}

\begin{table*}[!t]
\caption{Numerical results for {\DeltaVAE}s trained on (non-binarized) MNIST. The values indicate mean and standard deviation over $10$ runs. The columns represent the (data-averaged) log-likelihood estimate (LL), Evidence Lower Bound (ELBO), KL-divergence (KL) and mean squared error (MSE).}
\label{sample-table}
\vskip 0.15in
\begin{center}
\begin{small}
\begin{sc}
\begin{tabular}{ccccc}
\toprule
Manifold & LL & ELBO  & KL & MSE ($\times10^{-2}$)\\
\midrule
$S^2$ & -738.76$\pm$0.08 & -739.25$\pm$0.09 & 5.16$\pm$0.13 & 3.48$\pm$0.05
\\
Embedded Torus & -738.58$\pm$0.08 & -740.97$\pm$0.10 & 6.57$\pm$0.01 & 3.55$\pm$0.02\\

Flat Torus & -738.97$\pm$0.08 & -741.37$\pm$0.11 & 6.42$\pm$0.00 & 3.70$\pm$0.02
\\
$\mathbb{R}\mathbb{P}^3$ & -738.81$\pm$0.02 & -739.32$\pm$0.03 & 5.65$\pm$0.08 & 3.37$\pm$0.02
\\
$\mathbb{R}\mathbb{P}^2$ & -740.17$\pm$0.35 & -741.19$\pm$0.53 & 3.15$\pm$0.56 & 4.49$\pm$0.02
\\
$\mathbb{R}^3$ & -738.27$\pm$0.03 & -738.85$\pm$0.04 & 5.74$\pm$0.05 & 3.23$\pm$0.02
\\
$\mathbb{R}^2$ & -738.83$\pm$0.08 & -739.35$\pm$0.11 & 4.95$\pm$0.04 & 3.56$\pm$0.04\\

\bottomrule
\end{tabular}
\end{sc}
\end{small}
\end{center}
\vskip -0.1in
\end{table*}

\begin{table*}[!t]
\caption{Numerical results for {\DeltaVAE}s trained on a simple picture consisting of the lowest non-trivial Fourier components. The values indicate mean and standard deviation over $10$ runs. The columns represent the (data-averaged) log-likelihood estimate (LL), Evidence Lower Bound (ELBO), KL-divergence (KL) and mean squared error (MSE).}
\label{tab:fourier_components}
\vskip 0.15in
\begin{center}
\begin{small}
\begin{sc}
\begin{tabular}{ccccc}
\toprule
Manifold & LL & ELBO  & KL & MSE ($\times10^{-2}$)\\
\midrule
$S^2$ & -3779.75$\pm$2.25 & -3825.11$\pm$ 4.21 & 3.45$\pm$0.01 & 2.82$\pm$0.21\\
Embedded Torus & -3787.34$\pm$11.8 & -3809.33$\pm$23.0 & 11.2$\pm$1.20 & 1.85$\pm$1.18\\
Flat Torus & -3773.99$\pm$1.70 & -3813.00$\pm$5.16 &  6.42$\pm$0.00 & 0.90$\pm$0.25\\
$\mathbb{R}\mathbb{P}^3$ & -3774.61$\pm$0.53 & -3821.25$\pm$2.44 & 3.70$\pm$0.00 & 2.62$\pm$0.12\\
$\mathbb{R}\mathbb{P}^2$ & -3789.13$\pm$8.36 & -3850.19$\pm$33.7 &  3.92$\pm$1.97 & 4.02$\pm$1.74\\
$\mathbb{R}^3$ & -3779.67$\pm$4.06 & -3783.03$\pm$5.71 & 9.73$\pm$0.45 & 0.46$\pm$0.26\\
$\mathbb{R}^2$ & -3785.64$\pm$5.26 & -3789.60$\pm$6.19 & 8.26$\pm$0.53 & 0.85$\pm$0.28\\

\bottomrule
\end{tabular}
\end{sc}
\end{small}
\end{center}
\vskip -0.1in
\end{table*}

\subsubsection*{Numerical approximation and integration}

For some manifolds such as spheres or flat tori, exact solutions to the heat kernel are available, usually in terms of infinite series and special functions. For other, small-dimensional, Riemannian manifolds, the heat kernel may be accurately computed numerically. In both cases, the integration in the KL-term may be performed numerically.
If the dimensionality of the underlying manifold gets to large, one may have to resort to Monte Carlo approximation of the integral.

\section{Experiments}

We have implemented {\DeltaVAE}s with latent spaces of $d$-dimensional spheres, a flat two-dimensional torus, a torus embedded in $\mathbb{R}^3$, the $SO(3)$ and real projective spaces $\mathbb{RP}^d$. 

For all our experiments we used multi-layer perceptrons for the encoder and decoder with three and two hidden layers respectively. Recall that the encoder needs to produce both a point $z$ on the manifold and a time $t$ for the transition kernel. These functions share all layers, except for the final step where we project, with the projection map $P$, from the last hidden layer to the manifold to get $z$, and use an output layer with a $\tanh$ activation function to obtain $t$.

The encoder and decoder are connected by a sampling layer, in which we approximate sampling from the transition kernel of Brownian motion according to the reparametrization trick described in Section \ref{se:reparametrization-rw}.

\subsection{{\DeltaVAE}s for MNIST}

We then trained  {\DeltaVAE}s on MNIST. We show the manifolds as latent space with encoded MNIST digits in Figs.~\ref{fig:sphere-torus-latent} and \ref{fig:mnist-latent-rest}. When MNIST is trained on different latent spaces, different adjacency structures between digits may become apparent, providing topological information.

The $SO(3)$ is isometric to a scaling of the $\mathbb{RP}^3$ (with natural choices of Riemannian metrics). Although we have implemented an embedding and projection map for this embedding for the $SO(3)$ directly based on an SVD decomposition to find the nearest orthogonal matrix, training on the $\mathbb{RP}^3$ with the following trick was faster and we only present these results.

For training the projective spaces, we used an additional trick, where instead of embedding $\mathbb{RP}^d$ in a Euclidean space, we embed $S^d$ in $\mathbb{R}^{d+1}$, and make the decoder neural network \emph{even} by construction (i.e. the decoder applied to a point $s$ on the sphere equals the decoder applied to a point $-s$). Then, an encoder and decoder to and from the $\mathbb{RP}^3$ are defined implicitly. However, it must be noted that this setup does not allow for a homeomorphic encoding (because $\mathbb{RP}^d$ does not embed in $S^d$).

The numerically computed ELBO, reconstruction error, KL-divergence and mean-squared-error are shown in Table~\ref{sample-table} together with the estimated log-likelihood for a test dataset of MNIST. 

\subsubsection*{Estimation of log-likelihood}
For the evaluation of the proposed methods we have estimated the log-likelihood of the test dataset according to the importance sampling presented in \cite{Burda2015}.  The approximate log-likelihood of datapoint $x$ is calculated by sampling $L$ latent variables $z^{(1)}, \dots, z^{(L)}$ according to the approximate posterior $\mathbb{Q}_Z^{\mathbf{t}(x),\mathbf{z}(x)}$. The estimated log-likelihood for datapoint $x$ is given by
\[
\log{p_X^{\boldsymbol{\beta}}(x)}\approx\log\left(\frac{1}{L}{\sum_{l = 1}^L\frac{p_X^{\boldsymbol{\beta}(z^{(l)})}(x)p_Z(z^{(l)})}{q_Z(\mathbf{t}(x);\mathbf{z}(x),z^{(l)})}}\right).
\]
The log-likelihood estimates presented in Table \ref{sample-table} are obtained with $L = 100$ samples for each datapoint, averaged over all datapoints.
 
\begin{figure}[!t]
    \begin{minipage}{0.49\columnwidth}
    \centering
    \includegraphics[width=\textwidth]{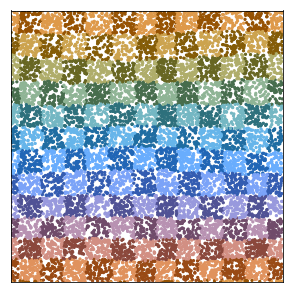}
    \subcaption[first caption.]{Latent space flat torus}\label{fig:fourier-1-flat-latent-1}
    \end{minipage}
    \begin{minipage}{0.49\columnwidth}
    \centering
    \includegraphics[width=\textwidth]{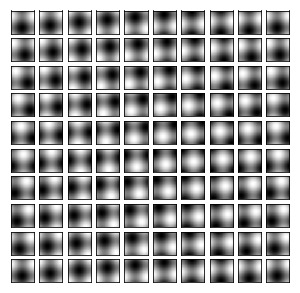}
    \subcaption[first caption.]{Reconstruction flat torus}\label{fig:fourier-1-flat-reconstruction}
    \end{minipage}
    \\
    \begin{minipage}{0.49\columnwidth}
    \centering
    \includegraphics[width=\textwidth]{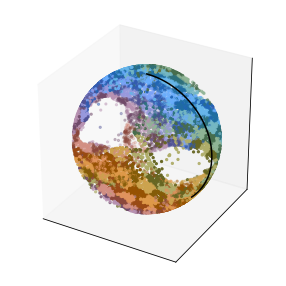}
    \subcaption[first caption.]{Latent space  sphere}\label{fig:sphere-latent-1}
    \end{minipage}
    \begin{minipage}{0.49\columnwidth}
    \centering
    \includegraphics[width=\textwidth]{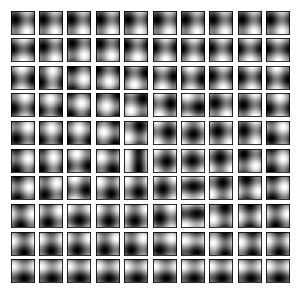}
    \subcaption[first caption.]{Reconstruction sphere}\label{fig:reconstruct-sphere}
    \end{minipage}
    \caption{Results of training {\DeltaVAE}s on a simple picture consisting only of the lowest non-trivial Fourier components. The figures on the left show the latent space manifolds of a flat torus and a sphere, with encoded images of translated original pictures, color-coded according to translation following the color scheme presented in Fig.~\ref{fig:checkers}. The figures on the right are reconstructions: they are placed on a grid in latent space and show the decoded images for the gridpoints. The reconstruction of the sphere is in spherical coordinates, with the left and right side of Fig.~\ref{fig:reconstruct-sphere} corresponding to the black arc on the sphere in Fig.~\ref{fig:sphere-latent-1}.}
\end{figure}

\subsection{Translations of periodic pictures}

To test whether a {\DeltaVAE} can capture topological properties, we trained it on synthetic datasets consisting of translations of the same periodic picture. 

Our input pictures were discretized to 64$\times$64 pixels, but to ease presentation, we discuss them below as if they were continuous. Note that with the choice of multi-layer perceptrons as encoder and decoder networks, the network has no information about which pixels are contiguous. 

\subsubsection*{Simple picture}

To illustrate the idea, we start with a very simple picture $\mathsf{pic}:[-\pi,\pi]^2 \to \mathbb{R}$, which only consists of the lowest Fourier components in each direction
\[
\mathsf{pic}(\theta,\phi) = \cos(\theta) + \cos(\phi).
\]
By translating the picture, i.e.~by considering different phases, we obtain a submanifold of the space of all pictures. When we interpret this space as $L^2([-\pi, \pi]^2)$, we get an embedding of the flat torus which is isometric up to a scaling. In other words, if we train the {\DeltaVAE} with a flat torus as latent space, we essentially try to find the identity map.

\begin{figure}
\begin{minipage}{0.49\columnwidth}
    \centering
    \includegraphics[width=\textwidth]{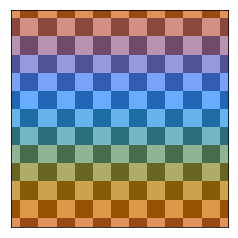}
\end{minipage}
\begin{minipage}{0.49\columnwidth}
    \caption{
    Colors used to color encoded pictures in latent space. The horizontal direction represents horizontal translation, the vertical direction represents vertical translation of the original, periodic picture. The boundary conditions are periodic.}
    \label{fig:checkers}
\end{minipage}
\end{figure}

Figs.~\ref{fig:fourier-1-flat-latent-1} and \ref{fig:fourier-1-flat-reconstruction} illustrate that for this simple case, the {\DeltaVAE} with a \emph{flat torus} as latent space indeed captures the translation of the picture as a latent variable. 
The fact that Fig.~\ref{fig:fourier-1-flat-latent-1} is practically a reflection and translation of the legend in Fig.~\ref{fig:checkers}, shows that there is an almost isometric correspondence between the translation of the original picture and the encoded latent variable. 

Although the figure does not provide a proof that the embedding is homeomorphic, it is in principle possible to give a guarantee that the encoder map is surjective (more precisely, that it has degree $1$ or $-1$).

We contrast this to when we train the same dataset on a {\DeltaVAE} with a sphere as a latent space. Fig.~\ref{fig:sphere-latent-1} displays typical results, showing that large parts of the sphere are not covered.

We present further numerical results in Table~\ref{tab:fourier_components}. We note that of all the closed manifolds, the MSE is by far the lowest for the flat torus. The KL divergence is lower for the positively-curved spaces $S^2$, $\mathbb{RP}^2$ and $\mathbb{RP}^3$, but the MSE for those is about a factor three higher. For $\mathbb{R}^3$ as latent space the MSE is even lower, but it is hard to compare because the KL term is not directly comparable, and it is the combination of the two that is optimized.

\subsubsection*{More complicated pictures}

\begin{figure}
    \begin{minipage}{0.49\columnwidth}
    \centering
    \includegraphics[width=\textwidth]{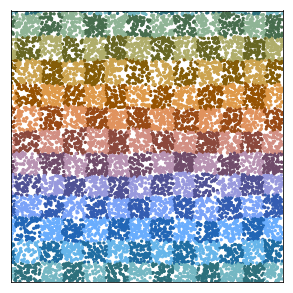}
    \end{minipage}
    \begin{minipage}{0.49\columnwidth}
    \centering
    \includegraphics[width=\textwidth]{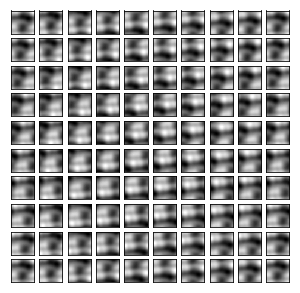}
    \end{minipage}
    \caption{Latent space and reconstruction for a {\DeltaVAE} with a flat torus as latent space, trained on a picture consisting of several Fourier components.}
    \label{fig:multiple-components}
\end{figure}

As an exploratory analysis of the extent to which the {\DeltaVAE} can capture the topological structure, we trained on fixed datasets, each consisting of translations of a fixed random picture. These random pictures, viewed as functions $\mathsf{pic}:[-\pi, \pi]^2\to \mathbb{R}$, are created by randomly drawing complex Fourier coefficients $a_{k,\ell}$ from a Gaussian distribution:
\[
\mathsf{pic}(\theta, \phi) = \mathrm{Re}\left(\sum_{k,\ell = -N}^N \gamma^{|k|+|\ell| -1} a_{k,\ell} \exp(i k \theta + i \ell \phi )\right)
\]
where $\gamma \in (0,1]$ is a discounting factor.
The {\DeltaVAE} with a flat torus as latent space is also capable of capturing the translations when the picture is generate this way, as long as the higher Fourier components carry not too much weight, see Fig.~\ref{fig:multiple-components}.

\subsubsection*{Too complicated pictures}

\begin{figure}
    \begin{minipage}{0.49\columnwidth}
    \centering
    \includegraphics[width=\textwidth]{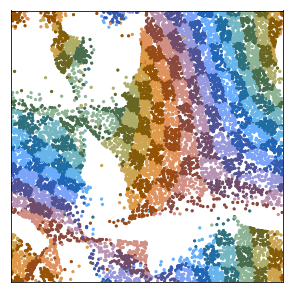}
    \end{minipage}
    \begin{minipage}{0.49\columnwidth}
    \centering
    \includegraphics[width=\textwidth]{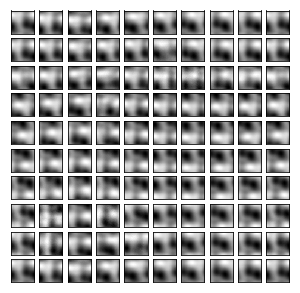}
    \end{minipage}
    \caption{Flat torus latent space and reconstruction for a {\DeltaVAE} trained on a picture with too much weight on higher Fourier components for the {\DeltaVAE} to capture the topological structure.}
    \label{fig:too-complicated}
\end{figure}

Generically, we see that when there is too much weight on the higher Fourier components in the picture ($N\gtrsim 10$ with $\gamma \gtrsim 0.3$), the {\DeltaVAE} is no longer capable of capturing the translations, see Fig.~\ref{fig:too-complicated}.

We only tested a very simple setup. It is likely that convolutional neural networks would significantly improve the performance in capturing the translations, since implicit information is added to the network about the underlying geometry of the pictures. 

Initial experiments with pretraining the network on simple Fourier images also show an increased success rate of capturing the topological structure.

\subsubsection*{Interesting patterns}

\begin{figure}
    \begin{minipage}{0.49\columnwidth}
    \centering
    \includegraphics[width=\textwidth]{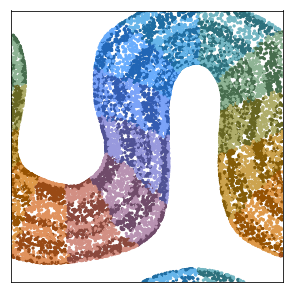}
    \end{minipage}
    \begin{minipage}{0.49\columnwidth}
    \centering
    \includegraphics[width=\textwidth]{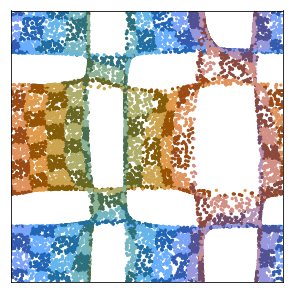}
    \end{minipage}
    
    \begin{minipage}{0.49\columnwidth}
    \centering
    \includegraphics[width=\textwidth]{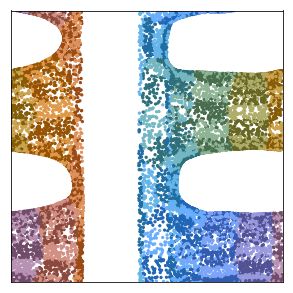}
    \end{minipage}
    \begin{minipage}{0.49\columnwidth}
    \centering
    \includegraphics[width=\textwidth]{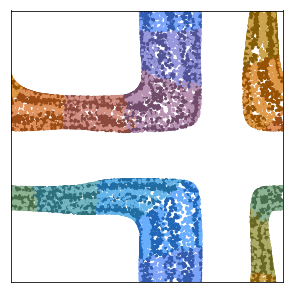}
    \end{minipage}
    \caption{Interesting patterns may occur when translations of a picture with not too much weight on higher Fourier components are encoded into the latent space of a flat torus.}
    \label{fig:interesting-patterns}
\end{figure}

For moderate weights on Fourier components, the data manifold is often mapped to the latent space in very interesting ways, see Fig.~\ref{fig:interesting-patterns}. The patterns that occur in this manner are very structured, suggesting that it may be possible to find underlying mathematical reasons for this structure. Moreover, it gives some indication that we may understand how the network arrives at such patterns, and how we may control them.

\section{Conclusion}

We developed and implemented Diffusion Variational Autoencoders, which allow for arbitrary manifolds as a latent space \footnote{\url{https://github.com/LuisArmandoPerez/DiffusionVAE}}. Our original motivation was to investigate to which extent VAEs find semantically meaningful latent variables, and more specifically, whether they can capture topological and geometrical structure in datasets. By allowing for an arbitrary manifold as a latent space,
{\DeltaVAE}s can remove obstructions to capturing such structure.

Indeed, our experiments with translations of periodic images show that a simple implementation of a $\Delta\!$VAE with a flat torus as latent space is capable of capturing topological properties. 

However, we have also observed that when we use periodic images with significant high Fourier components, it is more challenging for the {\DeltaVAE} to capture the topological structure. We will investigate further whether this can be resolved by better network design, by pretraining, different loss functions, or even more intensive extensions of the VAE algorithm. 

Our exploratory analysis shows that we can ask well-defined questions about whether a VAE can capture topological structure, and that such questions are amenable to a statistical approach. Moreover, the patterns observed in latent space in Fig.~\ref{fig:multiple-components} and Fig.~\ref{fig:interesting-patterns}, suggest that it may be possible to capture in mathematical theorems what the image of a data manifold in latent space would look like, when the networks are (almost) minimizing the loss. Combining these different perspectives, we may eventually develop more understanding on how to develop algorithms that capture semantically meaningful latent variables.

\bigskip

\bibliography{references2}
\bibliographystyle{icml2019}

\onecolumn
\appendix{}
\section{Experiments with binary MNIST}

The results for the the $\Delta$VAE trained on the binarized MNIST \cite{Salakhutdinov2008} for different manifolds are shown in Table \ref{tab:binarized_mnist}. We provide a comparison with the values obtained in \cite{Davidson2018} trained on a spherical latent space $S^2$ with a uniform prior. Additionally we present the results obtained in \cite{Figurnov2018} trained on a latent space consisting of two circular independent latent variables with a uniform prior, which can be directly compared to the $\Delta$VAE with a flat torus latent space.\\

The $\Delta$VAE achieves similar log-likelihood estimates with respect to the results on $\mathcal{S}^2$ from \cite{Davidson2018}. On the other hand, the results for the $\Delta$VAE trained on a flat torus have a lower log-likelihood compared to the results from \cite{Figurnov2018} (higher values are better).

\begin{table*}[h]
\caption{Numerical results for {\DeltaVAE}s trained on binarized MNIST. The values indicate mean and standard deviation over $10$ runs. The columns represent the (data-averaged) log-likelihood estimate (LL), Evidence Lower Bound (ELBO), KL-divergence (KL) and reconstruction error (RE). For comparison we present results for $S^2$ as reported by \cite{Davidson2018} and for the flat torus as reported by \cite{Figurnov2018}}.
\label{tab:binarized_mnist}
\vskip 0.15in
\begin{center}
\begin{small}
\begin{sc}
\begin{tabular}{ccccc}
\toprule
Manifold & LL & ELBO  & KL & RE\\
\midrule
$S^2$ & -132.20$\pm$0.39 & -134.67$\pm$0.47 & 7.23$\pm$0.05 & -127.44$\pm$0.47\\
Embedded Torus & -132.79$\pm$0.53 & -137.37$\pm$0.59 & 9.14$\pm$0.18 &  -128.23$\pm$0.67\\
Flat Torus & -131.73$\pm$0.69 & -139.97$\pm$0.78  & 12.91$\pm$0.08 & -127.07$\pm$0.81\\
$\mathbb{R}\mathbb{P}^3$ & -125.27$\pm$0.37 & -128.17$\pm$0.58 & 9.38$\pm$0.12 &  -118.79$\pm$0.60\\
$\mathbb{R}\mathbb{P}^2$ & -135.87$\pm$0.66 & -138.13$\pm$0.72 &  7.02$\pm$0.12 & -131.11$\pm$0.73 \\
$\mathbb{R}^3$ & -124.71$\pm$0.93 & -128.01$\pm$1.05 & 9.12$\pm$0.09 & -118.89$\pm$1.01\\
$\mathbb{R}^2$ & -134.17$\pm$0.53 & -136.61$\pm$0.64 &  7.05$\pm$0.06 & -129.56$\pm$0.63\\
$S^2$ \cite{Davidson2018} & -132.50$\pm$0.83 & -133.72$\pm$0.85 & 7.28$\pm$0.14 & -126.43$\pm$0.91
\\

Flat Torus\cite{Figurnov2018} & -127.60$\pm$0.40 &  -  & -  & - 
\\

\bottomrule
\end{tabular}
\end{sc}
\end{small}
\end{center}
\vskip -0.1in
\end{table*}

\section{Asymptotic expansion KL divergence}\label{app:asymptotic_expansion}
In this appendix, we derive a short-term asymptotic expansion of 
\begin{equation}
    \label{eq:entropy-integral}
\int_Z q_Z(t;z,w) \log q_Z(t;z,w) d w.
\end{equation}

We base the expansion on a short-term expansion of the heat kernel itself, also known as a parametrix expansion, cf. \cite{Berger-Spectre-1971}
\begin{equation}
\label{eq:parametrix-heat}
\begin{split}
q_Z(t; z, w) 
&:= \frac{1}{(2 \pi t)^{d/2}}
\exp\left(- \frac{r^2}{2t} \right) \left( u_0(z,w) + t u_1(z,w) + o(t) \right)\\
&=\frac{1}{(2 \pi t)^{d/2}}
\exp\left(- \frac{r^2}{2t} \right) u_0(z,w)
\left(1 + t \frac{u_1(z,w)}{u_0(z,w)} + o(t) \right),
\end{split}
\end{equation}
where $r$ is the distance between $z$ and $w$, and where we use the notation $o(t)$ for terms that go to zero faster than $t$.

Because the heat kernel decays exponentially, for calculating the asymptotic behavior in (\ref{eq:entropy-integral}), only the behavior of the function $u_j(z,w)$ for $z$ close to $w$ is relevant. We choose normal coordinates $y^i$ centered at $z$ (so $z$ corresponds to $y^i=0$, and $r^2 = |y|^2$), and Taylor expand the functions $u_j$ in terms of $y^i$. It is helpful to keep in mind as a rule of thumb, that a monomial of degree $k$ in $y^i$ corresponds to a factor of degree $k/2$ in $t$ in the final asymptotic expansion of the integral (\ref{eq:entropy-integral}).

We split the logarithm of $q_Z$ in four terms,
\begin{equation}
\label{eq:four-Js}
\log q_Z(t;z,w) = J_1(t;z,w) + J_2(t;z,w) + J_3(t;z,w) + J_4(t;z,w)
\end{equation}
where
\[
\begin{split}
J_1(t;z,w) &:= - \frac{d}{2} \log (2 \pi t)\\
J_2(t;z,w) &:= - \frac{r^2}{2t}\\
J_3(t;z,w) &:= \log u_0(z,w) \\
J_4(t;z,w) &:= \log\left(1 + t \frac{u_1(z,w)}{u_0(z,w)} + o(t) \right) \\
&=t \frac{u_1(z,w)}{u_0(z,w)} + o(t)
\end{split}
\]
where we used the Taylor expansion of the logarithm to get the second line of $J_4$.

Write 
\[
\theta(z,w) = \sqrt{\det (g_{ij}(w)) }
\]
where $g_{ij}$ are the coefficients of the metric in the normal coordinates $y^i$ centered at $z$. 

In the parametrix expansion, the function $u_0$ equals
\[
u_0(z,w) = \frac{1}{\sqrt{\theta(z,w)}}
\]
see p.~208 of \cite{Berger-Spectre-1971}.

Key are the following asymptotic expansions in normal coordinates $y^i$ centered at $z$,
\[
\theta(0, y) = 1 - \frac{1}{6} \mathrm{Ric}_{ij} y^i y^j + O (|y|^3),
\]
and
\begin{equation}
\label{eq:expansion-sqrt-theta}
\sqrt{\theta(0,y)} = 1 - \frac{1}{12} \mathrm{Ric}_{ij} y^i y^j + O(|y|^3).
\end{equation}
As a consequence, we have the following asymptotic expansion for $u_0$

\begin{equation}
\label{eq:expansion-u0}
u_0(0,y) = \frac{1}{\sqrt{\theta(0,y)}} = 1 + \frac{1}{12} \mathrm{Ric}_{ij} y^i y^j + O(|y|^3).
\end{equation}

Next, we use that the function $u_1$ is given by the following integral (see (E.III.1) in \cite{Berger-Spectre-1971}, but note that they have a different sign convention for the Laplacian, see formula (G.III.2) in their book, and that we use the stochastic normalization in the heat equation, which accounts for an extra factor of $\frac{1}{2}$)
\[
u_1(z,w) = \frac{1}{2} \theta^{-1/2}(z,w) \int_0^1 \theta^{1/2}\Big(z, \exp_z\big(\tau \exp_z^{-1}(w)\big)\Big) \Delta u_0 \Big(z, \exp_z\big(\tau \exp_z^{-1}(w)\big)\Big) d \tau
\]
where the Laplacian is taken in the second argument and where $\exp_z$ denotes the exponential map based at $z$.

Because the fraction $u_1(z,w)/u_0(z,w)$ gets multiplied by a factor $t$ in the parametrix expansion (\ref{eq:parametrix-heat}), we will later only need the zeroth order term of $u_1(z,w)/u_0(z,w)$ and $u_1(z,w)$. 
Since
\[
\Delta u_0 (0,y)= \frac{1}{12} 2 \mathrm{tr}(\mathrm{Ric}) + O(|y|) = \frac{1}{6} \mathsf{Sc} + O(|y|),
\]
we get from the integral formula that
\[
u_1(0,y) = \frac{1}{12} \mathsf{Sc} + O(|y|)
\]
and, by using (\ref{eq:expansion-u0}), that
\begin{equation}
\label{eq:expansion-u1-u0}
\frac{u_1(0,y)}{u_0(0,y)} = \frac{1}{12} \mathsf{Sc} + O(|y|).
\end{equation}

To compute the integral
\[
\int_Z q_Z(t; z, w) \log q_Z(t; z, w) dw
\]
we split the logarithm in four terms $J_i$ as in (\ref{eq:four-Js}). The first term gives
\[
\begin{split}
\int_Z q_Z(t;z,w) J_1(t;z,w) d w 
&= -\frac{d}{2} \int_Z q_Z(t; z, w) \log\left( 2\pi t \right) dw \\
&= -\frac{d}{2} \log(2 \pi t).
\end{split}
\]
The fourth term is also easy and gives
\[
\begin{split}
\int_Z q_Z(t; z, w) J_4(t;z,w) d w
&= \int_Z q_Z(t;z,w) \frac{1}{12} \mathsf{Sc}\, t \, d w + o(t)\\
&= \frac{1}{12} \mathsf{Sc}\, t + o(t).
\end{split}
\]
Now let us look at 
\[
\begin{split}
&\int_Z q_Z(t; z, w) J_2(t;z,w) d w
= - \int_Z q_Z(t;z,w) \frac{r^2}{2t} d w\\
&= - \int_{\mathbb{R}^d} \frac{1}{(2\pi t)^{d/2}} \exp \left( - \frac{|y|^2}{2t}\right)\frac{|y|^2}{2 t} 
\frac{1}{\sqrt{\theta (0,y)}}\left( 1 + t \frac{u_1(0,y)}{u_0(0,y)} + o(t)\right) \theta(0,y) d y\\
&= - \int_{\mathbb{R}^d} \frac{1}{(2\pi t)^{d/2}} \exp \left( - \frac{|y|^2}{2t}\right)\frac{|y|^2}{2 t} 
\left( 1 + t \frac{u_1(0,y)}{u_0(0,y)} + o(t)\right) \sqrt{\theta(0,y)} d y.
\end{split}
\]
We substitute the asymptotic behavior of $u_1(z,w)/u_0(z,w)$ from (\ref{eq:expansion-u1-u0}) and the asymptotic behavior of $\sqrt{\theta(0,y)}$ from (\ref{eq:expansion-sqrt-theta}),
\[
\begin{split}
&\left( 1+ t \frac{u_1(0,y)}{u_0(0,y)} + o(t) \right) \sqrt{\theta(0,y)} \\
&= \left( 1+ t \left(\frac{1}{12} \mathsf{Sc} + O(|y|) \right)+ o(t) \right) \left( 1 - \frac{1}{12} \mathrm{Ric}_{ij} y^i y^j + O(|y|^3)\right).
\end{split}
\]
We expand the factors and integrate. Note that the integral of an $O(|y|^k)$ term against the Gaussian measure can be, after integration, estimated by a term of $O(t^{k/2})$. We therefore find
\[
\begin{split}
&\int_Z q_Z(t; z, w) J_2(t;z,w) d w\\
&= - \int_{\mathbb{R}^d} \frac{1}{(2\pi t)^{d/2}} \exp \left( - \frac{|y|^2}{2t}\right)\frac{|y|^2}{2 t} 
\left( 1 + \frac{1}{12} \mathsf{Sc} \, t - \frac{1}{12} \mathrm{Ric}_{ij} \,  y^i y^j \right)  d y + o(t).
\end{split}
\]
All that is left to do is compute Gaussian integrals, which follow from the moments of a multidimensional Gaussian distribution with mean zero and a covariance of $t$ times the identity matrix. In particular, fixing $i$, we have
\[
\begin{split}
&\int_{\mathbb{R}^d} \frac{1}{(2\pi t)^{d/2}} \exp\left( -\frac{|y|^2}{2t}\right) |y|^2 (y^i)^2 dy\\
&= \int_{\mathbb{R}} \frac{1}{\sqrt{2\pi t}} \exp \left( - \frac{(y^i)^2}{2t} \right) (y^i)^4 dy  \\
&\quad + \sum_{j\neq i} \left(\int_{\mathbb{R}} \frac{1}{\sqrt{2\pi t}} \exp \left( - \frac{(y^j)^2}{2t} \right) (y^j)^2 dy^j\right)
\left(\int_{\mathbb{R}} \frac{1}{\sqrt{2\pi t}} \exp \left( - \frac{(y^i)^2}{2t} \right) (y^i)^2 dy^i\right)\\
&= 3 t^2 + (d-1) t^2.
\end{split}
\]

We find
\[
\begin{split}
&\int_Z q_Z(t; z, w) J_2(t;z,w) d w\\
&= - \frac{d}{2} - \frac{d}{24} \mathsf{Sc} \, t + \frac{1}{24}  t ( 3 + (d-1) ) \sum_{i=1}^d \mathrm{Ric}_{ii}\\
&= - \frac{d}{2} - \frac{d}{24} \mathsf{Sc} \, t + \frac{1}{24}\mathsf{Sc} \, t (d+2) \\
&= - \frac{d}{2} + \frac{1}{12} \mathsf{Sc} \, t.
\end{split}
\]
Finally, we consider
\[
\begin{split}
\int_Z q_Z(t; z, w) J_3(t;z, w) d w 
&= \int_Z q_Z(t;z,w) \log u_0(z,w) d w \\
&= \int_{\mathbb{R}^d} \frac{1}{(2 \pi t)^{d/2}} \exp \left( -\frac{|y|^2}{2t}\right)
\frac{1}{12} \mathrm{Ric}_{ij} y^i y^j d y + o(t)\\
&= \frac{1}{12} \mathsf{Sc} \, t + o(t).
\end{split} 
\]

If we add all contributions, we obtain
\[
\int_Z q_Z(t; z, w) \log q_z(t; z, w) d w
= - \frac{d}{2} \log (2\pi t) - \frac{d}{2} + \frac{1}{4} \mathsf{Sc} \, t + o(t).
\]

\end{document}